\documentclass[conference]{IEEEtran}
\usepackage{cite}
\usepackage{booktabs}
\usepackage{hyperref}
\usepackage{orcidlink}
\usepackage{tikz}
\usepackage{comment}
\usetikzlibrary{positioning,arrows.meta}
\usepackage{amsmath,amssymb,amsfonts}
\usepackage{algorithmic}
\usepackage{graphicx}
\usepackage{textcomp}
\usepackage{xcolor}
\usepackage{multirow}
\usepackage{bbm}
\usepackage[hang,flushmargin]{footmisc}
\DeclareMathOperator{\argmax}{argmax}
\def\BibTeX{{\rm B\kern-.05em{\sc i\kern-.025em b}\kern-.08em
    T\kern-.1667em\lower.7ex\hbox{E}\kern-.125emX}}
\begin{document}

\title{CRANE: Post-Hoc Counterfactual Retrieval for Multimodal Classifiers}

\author{\IEEEauthorblockN{1\textsuperscript{st} Franco Rugolon \orcidlink{0000-0002-7693-0576}}
\IEEEauthorblockA{\textit{Dept. of Computer and Systems Sciences} \\
\textit{Stockholm University}\\
Stockholm, Sweden \\
franco.rugolon@dsv.su.se}
\and
\IEEEauthorblockN{2\textsuperscript{nd} Ioanna Miliou \orcidlink{0000-0002-1357-1967}}
\IEEEauthorblockA{\textit{Dept. of Computer and Systems Sciences} \\
\textit{Stockholm University}\\
Stockholm, Sweden \\
ioanna.miliou@dsv.su.se}
\and
\IEEEauthorblockN{3\textsuperscript{rd} Panagiotis Papapetrou \orcidlink{0000-0002-4632-4815}}
\IEEEauthorblockA{\textit{Dept. of Computer and Systems Sciences} \\
\textit{Stockholm University}\\
Stockholm, Sweden \\
panagiotis@dsv.su.se}
}

\maketitle

\begin{abstract}
Counterfactual explanation methods have been introduced for unimodal settings, with dedicated frameworks for tabular, timeseries, text, and image data, yet no post-hoc method has been designed for classifiers that learn from multimodal inputs. Existing explainability methods for multimodal classifiers apply modality-specific attribution independently to each input, which fails to yield coherent explanations and does not transfer to arbitrary pre-trained multimodal architectures. To address these shortcomings, we present CRANE (Counterfactual Retrieval via Agnostic Nearest-neighbour Explanations), the first post-hoc, model-agnostic framework for nearest-neighbour multimodal counterfactual generation that retrieves real training examples from the target class as candidate explanations, ensuring plausibility by construction. % We further propose multimodal extensions of the standard counterfactual quality objectives, namely proximity, sparsity, and plausibility, grounded in a unified formulation that accommodates arbitrary combinations of input modalities. 
Across four heterogeneous datasets and three fusion strategies, we show that Multimodal Consensus Retrieval (MC-R), which intersects per-modality neighbour sets to identify instances jointly close to the query in every modality, emerges as the strongest multimodal strategy, achieving the best or joint-best outcome on three of four datasets while grounding every explanation in the joint training distribution. Unimodal strategies remain competitive when a single modality dominates the decision boundary.
\end{abstract}

\begin{IEEEkeywords}
Explainable artificial intelligence, Counterfactual explanations, Multimodal learning, Post-hoc explanation
\end{IEEEkeywords}

% ----------------------------------------------------------
%  SECTION 1 -- INTRODUCTION
% ----------------------------------------------------------
\section{Introduction}
\label{sec:intro}

Data science pipelines increasingly operate on heterogeneous inputs, combining structured records, free text, timeseries, and images within a single predictive model.
Clinical decision support systems fuse vital signs with structured patient demographics to predict outcomes, such as sepsis mortality~\cite{mondrejevski2024masicu} and cancer metastasis~\cite{rugolon2023workflow,rugolon2026multimodal,liu2025explainable}; content-moderation systems pair images and their captions to detect hateful memes~\cite{gomez2020exploring}; and fraud detectors combine free-text job postings with structured posting metadata to flag deceptive listings~\cite{akram2024online}.
As these multimodal architectures become standard practice, the need to explain their predictions grows. This need has driven a substantial body of work on explainable Artificial Intelligence (XAI)~\cite{arrieta2020xai,guidotti2019survey}.

Counterfactual explanations are among the most natural and actionable forms of post-hoc justification: a counterfactual $\mathbf{x}'$ reconfigures a sample~$\mathbf{x}$ so as to flip its prediction while staying close to~$\mathbf{x}$ and consistent with the data manifold. Falling within the example-based XAI methods, they answer the intuitive, often directly actionable question ``what would need to change in $\mathbf{x}$ for the model to decide otherwise?''.

%Counterfactual explanations are among the most natural forms of post-hoc justification: given a model's prediction on a sample~$\mathbf{x}$, a counterfactual $\mathbf{x}'$ reconfigures~$\mathbf{x}$, changing the model's prediction, while staying close to~$\mathbf{x}$ and consistent with the data manifold.
%sample $\mathbf{x}'$ that is as similar to~$\mathbf{x}$ as possible, while receiving a different prediction, 
%They answer the intuitive, often directly actionable question ``what would need to be changed in $\mathbf{x}$, so that the model decides otherwise?''.

\begin{figure}
    \centering
    \includegraphics[width=0.9\linewidth]{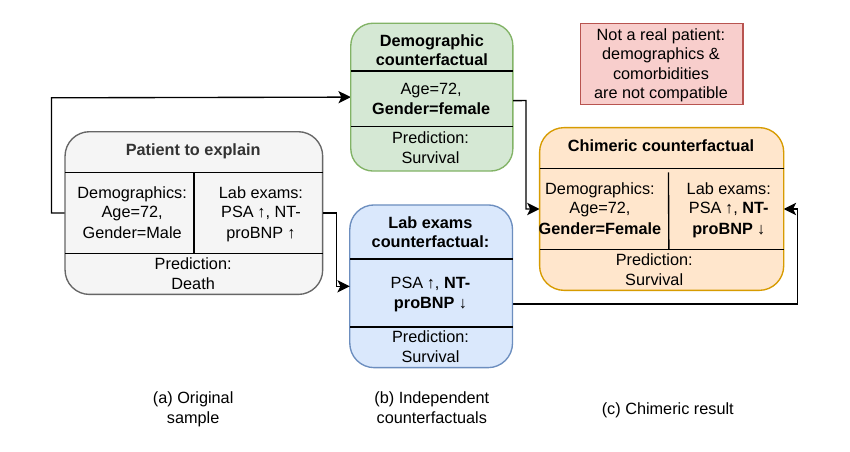}
    \caption{Independently generated per-modality counterfactuals can produce a chimeric explanation whose static features and timeseries trajectories are incoherent. Arrows denote rising~($\uparrow$) or falling~($\downarrow$) biomarkers.}
    \label{fig:chimeric}
\end{figure}

Well-performing counterfactual generators exist for individual modalities:  tabular~\cite{wachter2017counterfactual,mothilal2020explaining}, timeseries~\cite{ates2021counterfactual,delaney2021instancebased}, text~\cite{wu2021polyjuice}, and image~\cite{vermeire2022image}. These generators, however, assume a single, homogeneous input space. 
%onetheless, in many real-world applications, the underlying data are inherently multimodal, and consequently, classifiers leverage information from 
When a classifier consumes several modalities at once, applying these generators independently per modality fails to yield coherent explanations, as illustrated in Fig.~\ref{fig:chimeric}. For a multimodal mortality classifier over static demographics and laboratory timeseries, a tabular counterfactual generator may return a female survivor of the same age, while a timeseries generator may return signals that reflect decreasing N-terminal pro-B-type natriuretic peptide (NT-proBNP), a cardiac stress 
marker, but might not change the increasing trend in Prostate Specific Antigen (PSA), which is a clinical impossibility, as biological females do not produce PSA. This process would result in a chimeric counterfactual that would be locally plausible within each modality, yet globally incoherent and clinically meaningless when considered as a whole.

Retrieval-based methods avoid this problem by construction: rather than synthesising a new sample, they return a real training example from the target class. %, whose modalities were recorded together and are mutually coherent by definition.
The nearest-unlike-neighbour (NUN) literature exploits this in unimodal settings~\cite{brughmans2023nice,delaney2021instancebased,ates2021counterfactual}, but it has not been extended to the multimodal case.

In addition, when using a retrieval-based approach, different choices regarding the search space itself deserve consideration. One could search within a single modality's feature space, concatenate all modality embeddings and search jointly, or search in the multimodal model's latent representation. Each choice encodes a different notion of similarity, and how their relative merits compare in terms of the standard counterfactual quality objectives (prediction-flip rate, proximity, sparsity, and plausibility)~\cite{verma2024counterfactual}, and of the diversity of the generated set~\cite{mothilal2020explaining} remains to be explored.

%A second gap concerns the search space itself. Even granting a retrieval-based approach, one could search in a single modality's feature space, concatenate all modality embeddings and search jointly, or search in the model's own latent representation. Each choice encodes a different notion of similarity, and their relative merits in terms of standard counterfactual quality objectives (prediction-flip rate, proximity, sparsity, and plausibility)~\cite{verma2024counterfactual}, as well as diversity of the generated set~\cite{mothilal2020explaining}, have not been systematically studied.

We address these open questions with CRANE (Counterfactual Retrieval via Agnostic Nearest-neighbour Explanations), a post-hoc, model-agnostic framework for nearest-neighbour counterfactual generation for multimodal classifiers.
CRANE\footnote{The code is available on \href{https://github.com/FoxtrotRomeo/CRANE-cf}{GitHub}; the library can be installed with \texttt{pip install crane-cf}.} supports arbitrary combinations of tabular, timeseries, text, and image modalities through retrieval strategies spanning from unimodal search to multimodal fusion, and extends the standard counterfactual quality objectives (i.e., outcome, proximity, sparsity, and plausibility) to heterogeneous input spaces. Among the different strategies, Multimodal Consensus Retrieval (MC-R), which retrieves training instances jointly close to the query in each modality, emerges as the strongest multimodal strategy in four datasets spanning social media, recruitment, content moderation, and clinical prediction, while unimodal retrieval remains competitive when a single modality dominates the decision boundary.

\begin{comment}
The remainder of this paper is organised as follows: Section~\ref{sec:related} reviews related work on counterfactual explanations, multimodal learning, and explainability for multimodal models.
Section~\ref{sec:method} formalises the problem and describes the CRANE framework.
Section~\ref{sec:experiments} presents the experimental setup, and Section~\ref{sec:results} reports the results of our experiments.
Section~\ref{sec:conclusion} concludes with directions for future work.
\end{comment}
 
% ----------------------------------------------------------
%  SECTION 2 -- RELATED WORK
% ----------------------------------------------------------
\section{Related Work}
\label{sec:related}
 
%\subsection{Explainability in Machine Learning}
 
%The need to audit opaque Machine Learning (ML) models has given rise to a substantial body of work on explainable Artificial Intelligence (XAI)~\cite{arrieta2020xai,guidotti2019survey}, including feature-attribution methods, rule-extraction approaches, and example-based explanations.
%Counterfactuals fall within the example-based family, and have attracted particular attention because the changes they identify are directly actionable for practitioners. Despite this breadth, the vast majority of XAI methods assume a single, homogeneous input space: feature-attribution methods such as LIME~\cite{ribeiro2016lime} and SHAP~\cite{lundberg2017shap}, saliency-map methods, and attention-based explanations each target one data type. This limitation is sharpest for counterfactual methods, where the explanation must simultaneously specify what would change across all input spaces, not merely within one.
 
%\subsection{Counterfactual Explanation Methods}

Explainability methods for machine learning (ML) span a broad landscape, from feature-attribution methods~\cite{ribeiro2016lime,lundberg2017shap}, to saliency maps~\cite{selvaraju2017grad}, attention-based explanations~\cite{chefer2021transformer}, and example-based approaches, like counterfactuals, yet the vast majority assume a single, homogeneous input space. This limitation is sharpest for counterfactuals, where the explanation must simultaneously specify what would change across all input spaces, not merely within one. 
 
Counterfactual explanations were formalised in the context of ML by Wachter et al.~\cite{wachter2017counterfactual}, %who posed their generation as a constrained optimisation problem: find the sample closest to the query, under some distance metric, whose predicted class differs from the query's.
and have since grown into a rich research area. % was extended in subsequent work. 
For example, DiCE~\cite{mothilal2020explaining} generates a \emph{diverse} set of counterfactuals by adding a determinantal point process term to the objective, so that different explanations suggest different actionable paths, while Dandl et al.~\cite{dandl2020multi} proposed a multi-objective formulation that simultaneously optimises proximity, sparsity, plausibility, and actionability without collapsing them into a single weighted sum.
 
A distinct thread within the counterfactual literature avoids optimisation altogether by framing explanation as retrieval, where a real training example from the target class, the NUN, is used directly as the explanation or as a seed for further refinement.
%Rather than synthesising a new sample, retrieval-based methods return a real training example from the target class, the NUN, using it directly as the explanation or as a seed for further refinement.
Brughmans and Martens~\cite{brughmans2023nice} formalise this approach in NICE, which exploits the NUN to construct tabular counterfactuals that are guaranteed to exist in the observed data distribution.
The retrieval strategy carries an important structural guarantee: since the returned sample was observed in the training data, it is plausible by construction, without any explicit plausibility penalty in an objective function.

However, the methods reviewed above are designed exclusively for flat tabular or embedding-space inputs; the question of how to generate counterfactuals for classifiers that fuse multiple modalities simultaneously has received considerably less attention, as the following review of modality-specific approaches makes clear.

\section{Counterfactual Retrieval via Agnostic Nearest-neighbour Explanations}
\label{sec:method}

% ----------------------------------------------------------
\subsection{Problem Formulation}
\label{sec:formulation}
% ----------------------------------------------------------

We introduce a general formulation for multimodal counterfactual generation that is independent of any particular data type, classifier architecture, or application domain.

\smallskip
\noindent\textbf{Multimodal instance space.}
Let $M \geq 1$ be the number of modalities present in a dataset $\mathcal{S}$. For each modality $m \in \{1,\dots,M\}$, let $\mathcal{X}^m$ be the corresponding \emph{modality space}, encoding the representation format and domain constraints of that modality. The multimodal input space is
\begin{equation}
  \mathcal{X} \;:=\; \mathcal{X}^1 \times \cdots \times \mathcal{X}^M.
\end{equation}
A \emph{multimodal instance} is a tuple $x = (x^1, \dots, x^M) \in \mathcal{X}$, where $x^m \in \mathcal{X}^m$ is the component of $x$ in modality $m$.

In principle, $\mathcal{X}^m$ may represent any data type for which a suitable distance function $d_m$ and plausibility estimator $p_m$ can be defined. Without loss of generality, in this work we explore four modality types commonly encountered in heterogeneous data pipelines: static feature vectors (tabular), multivariate timeseries, token sequences (text), and two-dimensional pixel arrays (image). Adding additional modalities requires only specifying appropriate $d_m$ and $p_m$ functions for each one.

\smallskip
\noindent\textbf{Multimodal classifier.}
Let $\mathcal{Y}$ be a finite label set. A \emph{multimodal classifier} is a function $f: \mathcal{X} \to [0,1]^{|\mathcal{Y}|}$, where $f_y(x)$ is the score assigned to label $y \in \mathcal{Y}$ for instance $x$.
The predicted label is $\hat{y} := \arg\max_{y \in \mathcal{Y}} f_y(x)$. CRANE treats $f$ as a black box: only the prediction function is required, and no access to model internals, gradients, or training procedure is assumed.

\smallskip
\noindent\textbf{Multimodal counterfactual generation.}
Given a factual instance $x \in \mathcal{X}$ with predicted label $\hat{y} = f(x)$, a labelled training set $\mathcal{D} = \{(x_i, y_i)\}_{i=1}^{n}$, and a target label $y^\star \neq \hat{y}$, the goal is to find a counterfactual instance $x' \in \mathcal{X}$ such that $f(x') = y^\star$ and $x'$ is of high quality with respect to the multimodal equivalents to the standard counterfactual objectives, as defined in Section~\ref{sec:objectives}.

\subsubsection{Objectives}
\label{sec:objectives}
To evaluate the generated counterfactuals, we use four metrics commonly employed in the counterfactual explanation literature: \emph{outcome}, \emph{proximity}, \emph{sparsity}, and \emph{plausibility}. Each candidate counterfactual $\dot{x}$ is compared to the factual instance $x = (x_1, \dots, x_M)$. The latter three scores are computed per modality and aggregated using non-negative weights $w_m \geq 0$, $\sum_m w_m = 1$.
Every retrieval strategy returns an ordered list of such candidates; the best-scoring ones under the aggregated objectives are retained as counterfactual instances $x'$ as explained in Section~\ref{sec:formulation}.
All objectives are minimised; a value of zero denotes a perfect candidate.
To ensure that scores from modalities with different native scales are commensurable, every per-modality function is mapped to $[0,1]$ before aggregation, so the multimodal aggregate is guaranteed to lie in $[0,1]$ regardless of the number or type of modalities present.
In all experiments reported here we use uniform weights $w_m = 1/M$; dataset-specific weights may be substituted when prior knowledge about the relative importance of modalities is available. We define the four metrics as follows:

\paragraph{Outcome}
The first objective measures whether, and how decisively, a candidate attains the target label $y^\star$, using \emph{validity}, a binary indicator
\begin{equation}
  \mathrm{Validity}(\dot{x}) \;=\;
  \mathbbm{1}\!\left[\argmax_{y \in \mathcal{Y}} f_y(\dot{x}) = y^\star\right],
  \label{eq:validity}
\end{equation}
recording whether the prediction has flipped, and \emph{outcome}
\begin{equation}
  \Phi_{\text{out}}(\dot{x}) \;=\; 1 - f_{y^\star}(\dot{x}),
  \label{eq:outcome}
\end{equation}
a continuous deviation between the score the classifier assigns to the target class and the score it would assign to an unambiguous member of that class. $\Phi_{\text{out}}$ lies in $[0,1]$ and is minimised at $0$ when the classifier has full confidence in $y^\star$. 

\paragraph{Proximity}
Proximity measures how much the candidate deviates from the factual instance:
\begin{equation}
    \Phi_{\text{prox}}(x, \dot{x}) \;=\; \sum_{m=1}^{M} w_m\, \tilde{d}_m(x^m, \dot{x}^m),
  \label{eq:proximity}
\end{equation}
where $\tilde{d}_m \in [0,1]$ is a modality-specific normalised distance.
For tabular and timeseries modalities, the raw distance is the mean absolute error between $x^m$ and $\dot{x}^m$; it is normalised by converting it to a percentile rank against a reference distribution of pairwise distances computed from all cross-class training pairs, so that a score of $0$ indicates that the counterfactual is closer than all reference pairs and a score of $1$ indicates that it is farther.
For text and image modalities, $\tilde{d}_m$ is the normalised cosine distance, i.e.,
\begin{equation}
    \tilde{d}_m(x^m, \dot{x}^m) \;=\; \frac{1 - \cos\!\left(\phi_m(x^m),\, \phi_m(\dot{x}^m)\right)}{2},
  \label{eq:cosine_dist}
\end{equation}
where $\phi_m$ is the modality-specific encoder. Note that $\tilde{d}_m(\cdot)\in[0, 1]$, %, since cosine similarity takes values in $[-1, 1]$, 
ensuring comparability with the percentile-normalised distances used for tabular and timeseries modalities.

\paragraph{Sparsity}
Sparsity measures how many components of the factual instance must be altered
to yield the counterfactual:
\begin{equation}
    \Phi_{\text{spar}}(x, \dot{x}) \;=\; \sum_{m=1}^{M} w_m\, s_m(x^m, \dot{x}^m),
  \label{eq:sparsity}
\end{equation}
where $s_m \in [0,1]$ is a modality-specific normalised change count.
For the tabular modality, $s_m$ is the $\ell_0$ norm divided by the total
number of features $D$: $s_m = \|\dot{x}^m - x^m\|_0 / D$, so that a value of 1 indicates every feature has changed.
For the timeseries modality with $C$ channels, $s_m$ is computed per channel as the fraction of
timesteps at which the absolute difference exceeds a threshold
$\tau_c \geq 0$ (default $\tau_c = 0$, so any nonzero change is counted); these per-channel fractions are averaged over all $C$ channels, giving
$s_m \in [0,1]$.
For the text modality, $s_m$ is the token-level Levenshtein distance between
$x^m$ and $\dot{x}^m$, divided by $\max(1, |x^m|)$ where $|x^m|$ is the
number of tokens in the factual instance, and capped at $1.0$.
For the image modality, $s_m$ is the fraction of pixel elements (across all
channels, after normalising integer images to $[0,1]$) whose absolute
difference exceeds a threshold (default $0$).

\paragraph{Plausibility}
It measures how typical the candidate counterfactual is relative to the target-class training distribution:
\begin{equation}
    \Phi_{\text{plaus}}(\dot{x}) \;=\; \sum_{m=1}^{M} w_m\, p_m(\dot{x}^m),
  \label{eq:plausibility}
\end{equation}
where $p_m \in [0,1]$ is a normalised outlier score for modality $m$. For each modality, a Local Outlier Factor (LOF) model with $k = 20$ neighbours is fitted in novelty-detection mode on the target-class portion of the training set; timeseries are flattened to a vector of length $T \times C$ before fitting. The raw outlierness of a candidate $\dot{x}^m$ is $-\texttt{score\_samples}(\dot{x}^m)$, so higher values indicate a greater deviation from the training distribution.
This raw score is then mapped to $[0,1]$ via its percentile rank against the raw scores of all target-class training points:
\begin{equation}
    p_m(\dot{x}^m) \;=\; \frac{1}{n}\sum_{i=1}^{n}
    \mathbf{1}\!\left[\text{raw}(x_i^m) \leq \text{raw}(\dot{x}^m)\right],
  \label{eq:lof_percentile}
\end{equation}
where the sum is over the $n$ target-class training instances. A score of $0$ indicates that the candidate is at least as inlier-like as every target-class training point; a score of $1$ indicates that it is a stronger outlier than every target-class training point.

\smallskip
\noindent\textbf{On the plausibility of retrieved counterfactuals.}
A key property of retrieval-based methods is that whenever $k = 1$, the returned counterfactual is a real training instance from the target class and is therefore plausible by construction: it lies on the empirical data manifold, and all its modalities were observed together in a single real-world instance, preserving cross-modal coherence. When $k>1$, CRANE always returns the single nearest neighbour as its first candidate; the $j$-th subsequent candidate $(j = 2, \dots, k)$ is the component-wise median over the $j$ nearest neighbours, yielding a sequence of increasingly smoothed prototypes. In general, none of these median candidates corresponds to any observed training instance. The plausibility objective in Eq.~\eqref{eq:plausibility} quantifies plausibility for all candidates uniformly.

% ----------------------------------------------------------
\subsection{CRANE Retrieval Strategies}
\label{sec:crane}
% ----------------------------------------------------------

CRANE constructs counterfactual candidates by searching the training set $\mathcal{D}$ for the $k$ instances closest to $x$ among those labelled $y^\star$, and assembling a candidate from those neighbours. The key design choice is \emph{in which space} the nearest-neighbour search is performed. CRANE provides a family of retrieval strategies, ranging from unimodal search to multimodal fusion. All strategies share a common interface: they take $(x, y^\star, k)$ as input and return an ordered list of candidate counterfactuals. All treat the classifier $f$ as a black box. The different strategies are described below:

\subsubsection{Unimodal Retrieval}

CRANE provides four unimodal retrieval strategies — \textsc{TabularNN}, \textsc{TimeSeriesNN}, \textsc{TextNN}, and \textsc{ImageNN} — each ranking target-class training instances by distance in a single modality's feature or embedding space. Let $\mathcal{D}_{y^\star} = \{x_i \mid y_i = y^\star\}$ denote the target-class training instances. Given a modality-specific distance $d_m$, the $k$ nearest neighbours are
\begin{equation}
  N_m^{(k)}(x) \;=\; \operatorname{kNN}_{y^\star}\!\left(x^m;\, d_m\right),
  \label{eq:knn}
\end{equation}
the $k$ elements of $\mathcal{D}_{y^\star}$ with smallest $d_m(x^m, x_i^m)$.
For text and image modalities, $d_m$ operates in encoder embedding space: $d_m(a,b) = 1 - \cos\!\left(\phi_m(a),\,\phi_m(b)\right)$, where $\phi_m$ is a configurable pre-trained encoder treated as a hyperparameter of the distance function. In the experiments reported here we evaluate BERT-family models (multilingual E5, MiniLM), TF-IDF, and Word2Vec for text; ResNet-50, EfficientNet-B0, ViT-B/16, and CLIP ViT-B/32 for images. All four strategies share the same retrieval interface, but the rule for constructing the $j$-th candidate ($j = 1,\dots,k$) from the ranked neighbours depends on whether the modality is structured or unstructured, and this choice has direct consequences for how quality metrics behave as $k$ increases.

\smallskip
\noindent\textbf{Structured modalities (tabular, timeseries).}
Feature values are defined on a common continuous scale, so a component-wise median over $j$ neighbours is a meaningful summary. The $j$-th candidate prototype is the progressive median over the $j$ nearest neighbours:
\begin{equation}
  \pi_m^{(j)}(x) \;=\; \operatorname*{median}_{x_i \,\in\, N_m^{(j)}(x)} x_i^m.
  \label{eq:structured_proto}
\end{equation}
At $j = 1$ this returns a real training instance. For $j > 1$ the prototype is a smoothed aggregate that may not correspond to any observed row, but it remains close to the centre of the local target-class neighbourhood. As a result, proximity and plausibility tend to improve modestly as $j$ increases, since the progressive median averages out outlying feature values. \textsc{TabularNN} uses Euclidean or Manhattan distance; \textsc{TimeSeriesNN} uses Euclidean, DTW, or LCSS.

\smallskip
\noindent\textbf{Unstructured modalities (text, image).}
No meaningful median exists in the original representation space: the component-wise median of token sequences or pixel arrays is not a coherent text or image. CRANE therefore uses rank-matched selection: the $j$-th candidate takes the literal text or image from the $j$-th nearest neighbour in embedding space,
\begin{equation}
  \pi_m^{(j)}(x) \;=\; x_{(j)}^m,
  \label{eq:unstructured_proto}
\end{equation}
where $x_{(j)}$ denotes the $j$-th nearest neighbour under $d_m$. Therefore, every candidate is a real observed training instance in any $j$, but as $j$ increases, the selected instances are drawn from progressively more distant regions of the embedding space. Proximity and sparsity objectives for text and image are measured in embedding space, since recovering an unambiguous token sequence or pixel array from an encoder representation is not possible, in general.

\subsubsection{Early-Fusion Nearest Neighbour (EF-NN)}

Early-fusion retrieval forms a joint representation by concatenating the embeddings of all modalities into a single vector:
\begin{equation}
  \tilde{x} \;=\; \mathrm{concat}\!\left(\phi_1(x^1),\,\dots,\,\phi_M(x^M)\right),
\end{equation}
and retrieves the $k$ nearest neighbours in this joint space using a distance $\rho$ defined on the concatenated vector (e.g., cosine or Euclidean).
The prototype is assembled from the jointly-ranked neighbours using the modality-type materialisation rules defined in Section~\ref{sec:crane} (Eqs.~\ref{eq:structured_proto}--\ref{eq:unstructured_proto}). The only difference from a unimodal strategy is the ranking step: neighbours are ordered by distance in the concatenated space $\tilde{x}$ rather than in any single modality's space.

\subsubsection{Intermediate-Fusion Nearest Neighbour (IF-NN)}

Intermediate-fusion retrieval performs neighbour search in the latent representation learned by the classifier. Let $\lambda : \mathcal{X} \to \mathbb{R}^d$ denote the penultimate-layer activation of the multimodal classifier $f$. Distances are computed as
\begin{equation}
  d_{\mathrm{int}}(x, x_i) \;=\; \|\lambda(x) - \lambda(x_i)\|_2,
\end{equation}
and the $j$-th candidate is assembled from the $j$ nearest neighbours in this representation. Because $\lambda(x)$ is a single continuous vector regardless of the original modality types, candidate materialisation always follows the structured rule: the $j$-th candidate is the component-wise median over the latent-space-ranked neighbours, mapped back to input space by retrieving the corresponding complete training instances and applying Eqs.~\ref{eq:structured_proto}--\ref{eq:unstructured_proto} per modality.
By operating in the model's own embedding space, IF-NN retrieves instances that the classifier itself considers similar to $x$. However, unlike all other CRANE strategies, which operate solely through a prediction function and are fully model-agnostic, this one requires the explained model to be a deep neural network with accessible intermediate activations; it is not applicable to non-differentiable models.

\subsubsection{Multimodal Consensus Retrieval (MC-R)}

MC-R retrieves instances that are simultaneously close to $x$ across all modalities by intersecting per-modality neighbour sets. For each modality $m$, the top-$k$ nearest target-class neighbours are retrieved independently:
\begin{equation}
  N_m^{(k)}(x) \;=\; \operatorname{kNN}_{y^\star}\!\left(x^m;\, d_m\right).
\end{equation}
The consensus set is the intersection of these sets:
\begin{equation}
  \mathcal{S}(x) \;=\; \bigcap_{m=1}^{M} N_m^{(k)}(x).
\end{equation}
$k$ is chosen large enough to ensure $\mathcal{S}(x) \neq \emptyset$. The $j$-th candidate is then assembled from the $j$ nearest instances in $\mathcal{S}(x)$, ranked by the summed per-modality distance, using the same materialisation rules (Eqs.~\ref{eq:structured_proto}--\ref{eq:unstructured_proto}). MC-R is the most conservative multimodal strategy: candidates are drawn from the intersection of per-modality neighbourhoods and are therefore jointly close to $x$ in every modality simultaneously. The first candidate (when $|\mathcal{S}(x)| \geq 1$) is always a real training instance.

\subsubsection{Modality-wise Prototype Synthesis (MPS)}

MPS synthesises a candidate by combining independent per-modality prototypes. For each modality $m$, the $k$ nearest neighbours are retrieved in that modality's own feature or embedding space, as in Eq.~\ref{eq:knn}.  The per-modality prototype $\pi_m^{(j)}(x)$ for the $j$-th candidate is computed following the modality-type rules in Eqs.~\ref{eq:structured_proto}--\ref{eq:unstructured_proto}. The $j$-th synthetic counterfactual is assembled by concatenating the per-modality prototypes:
\begin{equation}
\dot{x}_{\mathrm{MPS}} \;=\; \bigl(\pi_1^{(j)}(x),\,\dots,\,\pi_M^{(j)}(x)\bigr).
\end{equation}
Unlike MC-R, MPS does not require any training instance to rank highly in all modalities simultaneously, so it can produce multimodal candidates even when no single instance is jointly close to $x$ across all modalities. However, because each modality is sourced from a different nearest neighbour, MPS candidates are synthetic composites that do not in general correspond to any observed training instance, even at $j = 1$. Consequently, MPS does not carry the structural plausibility guarantees that hold for MC-R and the unimodal retrievers.

\section{Experiments}
\label{sec:experiments}

% ----------------------------------------------------------
\subsection{Datasets}
\label{sec:datasets}
% ----------------------------------------------------------

We evaluate CRANE on four publicly available datasets spanning diverse application domains, modality combinations, and class structures. A summary is given in Table~\ref{tab:datasets}. %All non-Sepsis datasets are split 80/20 with stratified sampling; macro-F1 is the primary metric throughout.

\smallskip
\noindent\textbf{Sepsis (in-hospital mortality).}
The Sepsis dataset is extracted from MIMIC-III~\cite{johnson2016mimic} following the procedure described in~\cite{mondrejevski2024masicu}, which identifies ICU admissions with a confirmed sepsis diagnosis. Each admission is represented as 24 hourly timesteps over 100 features, partitioned into a timeseries modality of 53 time-varying clinical variables (vital signs and laboratory values) and a tabular modality of 47 static features (demographic and admission data extracted from the first timestep). The classification task is binary in-hospital mortality prediction. The dataset is strongly imbalanced, with a mortality rate of approximately 7.4\%. Evaluation follows 5-fold stratified cross-validation; results are pooled across folds and reported as F1 score as the primary metric, and AUC.

\noindent\textbf{Long COVID Tweets (emotion classification).}
This dataset~%\cite{santarabantoosoo2022longcovid} 
\footnote{\url{https://huggingface.co/datasets/Santarabantoosoo/italian_long_covid_tweets}}
consists of Italian tweets discussing long COVID, annotated with four emotion labels drawn from the FEEL-IT taxonomy~\cite{bianchi2021feel}: fear, anger, sadness, and joy.
Each instance comprises a raw tweet text field and a tabular modality encoding engagement metadata (likes, retweets, replies, quote counts) and structural tweet features (hashtag and URL counts). The class distribution is imbalanced, with fear (41.6\%) and anger (32.9\%) dominating and joy (8.8\%) forming the minority class.

\smallskip
\noindent\textbf{Hateful Memes (hate speech detection).}
The Hateful Memes dataset~\cite{kiela2020hateful} is a binary classification benchmark for detecting hate speech in internet memes. Each instance consists of a text caption and a 224$\times$224 RGB image. The class distribution is moderately imbalanced, with 63.7\% non-hateful and 36.3\% hateful instances. %This is the only dataset in our benchmark where the image modality is present.

\smallskip
\noindent\textbf{Real or Fake Jobs (fraud detection).}
This dataset~\cite{vidros2017automatic} contains job postings labelled as real or fraudulent. Each instance combines a tabular modality (structured fields such as salary range, employment type, and required experience) with three independent text fields: job description, company profile, and requirements. The task is strongly imbalanced, with fraudulent postings accounting for only 4.8\% of the data.

\begin{table}[t]
\centering
\caption{Summary of datasets used in the experiments.}
\label{tab:datasets}
\small
\resizebox{0.9\linewidth}{!}{
\begin{tabular}{lllrr}
\toprule
Dataset & Modalities & Task & Train & Test \\
\midrule
Sepsis            & TS + Tab                   & Binary   & 2,085/fold & 521/fold  \\
Long COVID Tweets & Text + Tab                 & 4-class  & 20,121           & 5,031  \\
Hateful Memes     & Text + Image               & Binary   & 7,632            & 1,908  \\
Real or Fake Jobs & Text + Tab        & Binary   & 14,304           & 3,576  \\
\bottomrule
\end{tabular}}
\end{table}

% ----------------------------------------------------------
\subsection{Experimental Setup}
\label{sec:setup}
% ----------------------------------------------------------

\subsubsection{Classifier Families and Fusion Strategies}

For each dataset, we train classifiers under three fusion strategies: early fusion, intermediate fusion, and late fusion, as defined in Section~\ref{sec:crane}. For each fusion strategy, we compare two classifier families: a deep family based on multi-layer perceptrons (MLP) with task-trained branches, and a non-deep family using tree-based ensembles (Random Forest and Gradient Boosting) or, for timeseries branches, the ROCKET transform with Logistic Regression~\cite{dempster2020rocket}. This design yields a range of classifier types that stress-tests CRANE's model-agnostic claim: the IF-NN retrieval strategy requires a deep neural classifier with accessible penultimate activations and is therefore evaluated only on deep models, while all other CRANE strategies are evaluated uniformly across both families.

Text is encoded using pre-trained transformer models: XLM-RoBERTa for Long COVID Tweets, DistilBERT for Real or Fake Jobs, and BERT or MiniLM for Hateful Memes. Images in Hateful Memes are encoded using ResNet-50, EfficientNet-B0, or ViT-B/16. For early-fusion classifiers, all encoders are frozen (no fine-tuning); for intermediate-fusion classifiers, the transformer backbone is fine-tuned end-to-end. %as part of the classifier training procedure.
Class imbalance is addressed via inverse-frequency weighting for all classifiers. For training we use a 15\% hold-out validation split for early stopping and hyperparameter tuning.

\subsubsection{CRANE Configuration}

Each CRANE retrieval strategy is configured with $k = 20$, returning 20 candidates per query. For MC-R, the per-modality search pool is set to $k_{\text{search}} = 200$ neighbours per modality prior to intersection, ensuring that the consensus set is non-empty for all queries in our datasets. For each dataset we run a systematic ablation over all valid combinations of encoder and distance metric per modality (as defined in Section~\ref{sec:crane}), yielding 6, 22, 16, and 84 combinations for Sepsis, Long COVID Tweets, Real or Fake Jobs, and Hateful Memes, respectively. The results are reported as averages across all ablation combinations.% for each dataset and retrieval strategy.

\begin{comment}
\begin{table}[t]
\centering
\caption{Number of encoder/distance combinations evaluated per dataset.}
\label{tab:ablation}
\small
\resizebox{0.48\textwidth}{!}{\begin{tabular}{lrl}
\toprule
Dataset & Combinations & Modality-specific variation \\
\midrule
Sepsis            & 6  & Tab $\times$ TS: $\{$Euc, Man$\} \times \{$DTW, Euc, LCSS$\}$ \\
Long COVID Tweets & 22 & Tab $\times$ Text: $2 \times 11$ metric/encoder combos \\
Real or Fake Jobs & 16 & Tab $\times$ Text: $2 \times 8$ metric/encoder combos \\
Hateful Memes     & 84 & Text $\times$ Image: $14 \times 6$ metric/encoder combos \\
\bottomrule
\end{tabular}}
\end{table}
\end{comment}

\subsubsection{CRANE Strategies per Dataset}

Not all retrieval strategies are applicable to every dataset and classifier combination. IF-NN is evaluated only on deep classifiers, as described in Section~\ref{sec:setup}; for datasets where a modality is absent (e.g., no image modality in Sepsis, Tweets, or Jobs), the corresponding unimodal strategy is not evaluated.

\begin{comment}
\begin{table}[t]
\centering
\caption{Applicability of CRANE retrieval strategies per dataset.
\checkmark~= evaluated; $-$~= not applicable (absent modality).}
\label{tab:applicability}
\small
\resizebox{0.48\textwidth}{!}{
\begin{tabular}{lcccccccc}
\toprule
Dataset & Tab & TS & Text & Img & EF & IF$^\dagger$ & MC-R & MPS \\
\midrule
Sepsis & \checkmark & \checkmark & $-$        & $-$        & \checkmark & \checkmark & \checkmark & \checkmark \\
Tweets & \checkmark & $-$        & \checkmark & $-$        & \checkmark & \checkmark & \checkmark & \checkmark \\
Memes  & $-$        & $-$        & \checkmark & \checkmark & \checkmark & \checkmark & \checkmark & \checkmark \\
Jobs   & \checkmark & $-$        & \checkmark & $-$        & \checkmark & \checkmark & \checkmark & \checkmark \\
\bottomrule
\end{tabular}}
\vspace{2pt}
\begin{minipage}{\linewidth}\footnotesize
Tab = TabularNN; TS = TimeSeriesNN; Text = TextNN; Img = ImageNN;
EF = EF-NN; IF = IF-NN; MC-R = Multimodal Consensus Retrieval;
MPS = Modality-wise Prototype Synthesis.
$^\dagger$ IF-NN evaluated on deep classifiers only; not applicable to RF, GBT, or ROCKET classifiers.
\end{minipage}
\end{table}
\end{comment}

% ----------------------------------------------------------
\subsection{Evaluation Protocol}
\label{sec:evaluation}
% ----------------------------------------------------------

We evaluate each retrieval strategy using the four quality objectives introduced in Section~\ref{sec:objectives}. All metrics are reported on a lower-is-better scale, where 0.0 denotes a perfect score on that objective.
%For Sepsis, all four metrics are averaged across the five cross-validation folds using pooled test queries. For the three single-split datasets, metrics are computed on the fixed held-out test set. 
All four metrics are averaged across the five cross-validation folds using pooled test queries. For each dataset and strategy, we report the mean and standard deviation across all ablation combinations, providing a measure of sensitivity to encoder and distance metric choice.
Counterfactual generation is performed on all test instances that the explained classifier predicts to belong to an undesired class; no subsampling is applied. For binary datasets, the target label $y^\star$ is the single opposite class. For Long COVID Tweets, where four emotion classes are present, we focus on the sadness$\to$joy direction, as these two classes represent the most semantically opposed poles in the dataset; the remaining class pairs (fear and anger) do not have a clearly defined opposite and are excluded from the counterfactual evaluation.

\section{Results}
\label{sec:results}
\subsection{Classifier Performance}
\label{sec:classifier_performance}

Table~\ref{tab:classifiers} reports macro-F1, accuracy, and AUROC for the best-performing classifier per fusion strategy and dataset, selected on the held-out set. Across datasets, intermediate fusion consistently yields the strongest or joint-strongest macro-F1, most notably on Long COVID Tweets (0.717 vs.\ 0.589 for early fusion) and Real or Fake Jobs (0.932 vs.\ 0.886 for early fusion). On Sepsis, late fusion achieves the highest AUROC (0.871). Performance on Hateful Memes is more modest across all strategies (macro-F1 0.669-0.713). These classifiers serve as the explained models in all subsequent counterfactual experiments.

\begin{table}[tb]
\centering
\caption{Best classifier per fusion strategy and dataset.}
\label{tab:classifiers}
\small
\resizebox{0.9\linewidth}{!}{
\begin{tabular}{lllrrr}
\toprule
Dataset & Strategy & Model & Macro F1 & Acc & AUROC \\
\midrule
\multirow{3}{*}{Sepsis}
  & Early        & Spatiotemporal Attention + GRU    & 0.691 & 0.920 & 0.800 \\
  & Intermediate & RF on 32-dim latents              & 0.746 & 0.946 & 0.833 \\
  & Late         & GRU + MLP branches                & 0.691 & 0.912 & 0.871 \\
\midrule
\multirow{3}{*}{\shortstack[l]{Long COVID\\Tweets}}
  & Early        & MLP on frozen RoBERTa             & 0.589 & 0.650 & 0.855 \\
  & Intermediate & Fine-tuned transformer + MLP      & 0.717 & 0.759 & 0.921 \\
  & Late         & TF-IDF + LR / GBT branches        & 0.604 & 0.658 & 0.844 \\
\midrule
\multirow{3}{*}{\shortstack[l]{Hateful\\Memes}}
  & Early        & GBT on frozen embeddings          & 0.713 & 0.740 & 0.777 \\
  & Intermediate & Fine-tuned transformer + MLP      & 0.698 & 0.721 & 0.748 \\
  & Late         & TF-IDF + LR / GBT branches        & 0.669 & 0.703 & 0.723 \\
\midrule
\multirow{3}{*}{\shortstack[l]{Real or\\Fake Jobs}}
  & Early        & RF on frozen embeddings           & 0.886 & 0.982 & 0.990 \\
  & Intermediate & Fine-tuned transformer + MLP      & 0.932 & 0.988 & 0.992 \\
  & Late         & TF-IDF + LR / GBT branches        & 0.922 & 0.986 & 0.988 \\
\bottomrule
\end{tabular}}
\end{table}

\subsection{Generator quality}
\label{sec:generator_quality}

\begin{figure}
    \centering
    \includegraphics[width=0.9\linewidth]{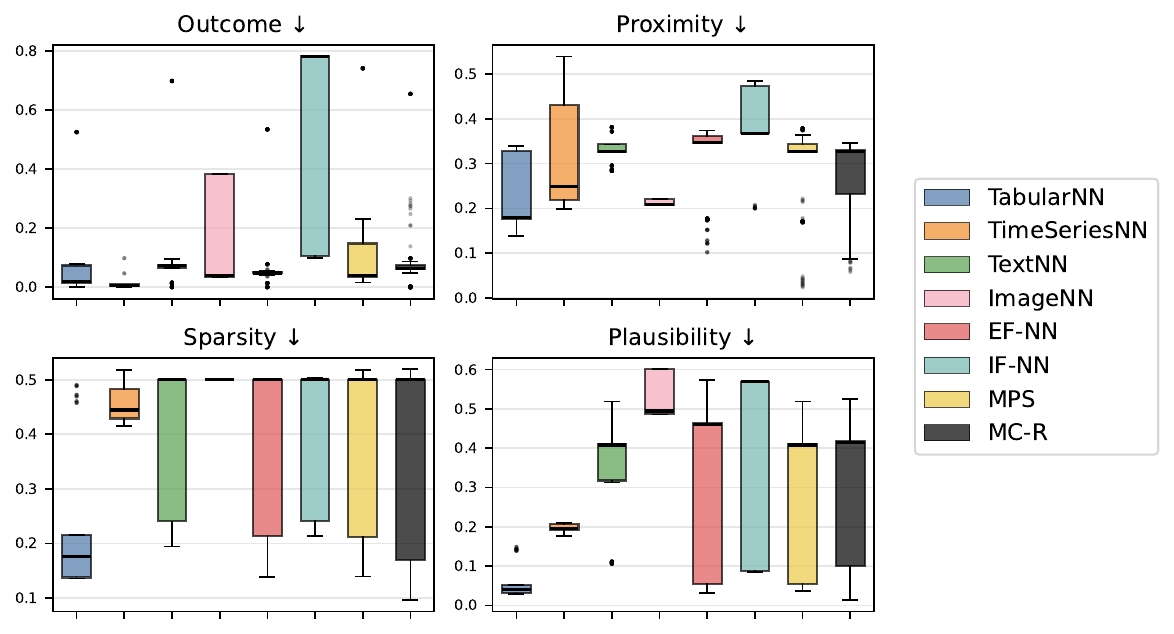}
    \caption{Generator quality across the full benchmark ($k = 20$). Each box shows the distribution over all encoder/distance-metric combinations, datasets, and fusion strategies. All metrics are lower-is-better.}
    \label{fig:global_boxplot}
\end{figure}

Unless otherwise stated, all results in the following sections are reported for $k = 20$; the effect of varying $k$ is analysed separately in Section~\ref{sec:k_ablation}.
Fig.~\ref{fig:global_boxplot} shows the distribution of each quality metric across all datasets and fusion strategies, with each box capturing variance across encoder and distance-metric combinations within that generator, while Table~\ref{tab:generator_perdataset} reports per-dataset results averaged across all three fusion strategies and encoder/distance-metric combinations.
 
%The most consistent finding across all datasets is the behaviour of IF-NN: with a mean outcome of 0.526, it fails to flip the classifier's prediction in more than half of all candidates on average, far exceeding every other generator. This holds across all four datasets and all three fusion strategies, independently of the choice of encoder or distance metric, as shown in Table~\ref{tab:generator_perdataset}.
%The IF-NN underperformance holds uniformly across all four settings (outcome 0.371 on Sepsis, 0.699 on Tweets, 0.816 on Memes, 0.113 on Jobs). 

IF-NN underperforms consistently across all four datasets compared to the other generators, with outcome values of 0.371 on Sepsis, 0.699 on Tweets, 0.816 on Memes, and 0.113 on Jobs. Consequently, it emerges as the weakest generator overall, with a mean outcome of 0.526 and a validity rate of only 46.3\%. %This trend remains unchanged across all three fusion strategies and is independent of the encoder or distance metric used, as shown in Table~\ref{tab:generator_perdataset}.
Among the remaining generators, EF-NN achieves the lowest mean outcome (0.079), making it the most reliable at producing valid counterfactuals across the benchmark. Unimodal generators achieve the best proximity and plausibility. 
 
On Real or Fake Jobs, MC-R achieves an outcome of 0.000 alongside the lowest sparsity (0.137) in the entire benchmark. On Sepsis, MPS achieves the best proximity (0.093) and plausibility (0.129), while TimeSeriesNN has the lowest outcome (0.021). On Long COVID Tweets, TabularNN achieves the best outcome (0.217), with engagement metadata proving more useful for retrieval than free-form tweet text. Finally, the constant sparsity of the Memes dataset (0.500) reflects a structural property: any two distinct training images are pixel-different everywhere after normalisation, so sparsity is fixed at 0.5 regardless of generator or $k$.
 
\begin{table}[ht]
\centering
\caption{Generator quality per dataset at $k = 20$, averaged across all three fusion strategies and encoder/distance-metric combinations (lower is
better; best result per metric/dataset in bold).}
\label{tab:generator_perdataset}
\small
\resizebox{0.88\linewidth}{!}{
\begin{tabular}{llcccc}
\toprule
Dataset & Generator & Outcome & Proximity & Sparsity & Plausibility \\
\midrule
\multirow{6}{*}{Sepsis}
 & TabularNN    & 0.056 & 0.158 & 0.474 & 0.143 \\
 & TimeSeriesNN & \textbf{0.021} & 0.317 & \textbf{0.457} & 0.196 \\
 & EF-NN        & 0.055 & \textbf{0.131} & 0.465 & 0.143 \\
 & IF-NN        & 0.371 & 0.203 & 0.488 & 0.151 \\
 & MC-R         & 0.163 & 0.119 & 0.476 & 0.242 \\
 & MPS          & 0.124 & 0.093 & 0.450 & \textbf{0.129} \\
\midrule
\multirow{6}{*}{Tweets}
 & TabularNN    & \textbf{0.217} & 0.350 & \textbf{0.142} & \textbf{0.052} \\
 & TextNN       & 0.286 & 0.449 & 0.202 & 0.325 \\
 & EF-NN        & 0.222 & \textbf{0.343} & 0.145 & 0.054 \\
 & IF-NN        & 0.699 & 0.570 & 0.219 & 0.166 \\
 & MC-R         & 0.300 & 0.356 & 0.172 & 0.089 \\
 & MPS          & 0.298 & 0.334 & 0.144 & 0.054 \\
\midrule
\multirow{6}{*}{Memes}
 & TextNN       & 0.070 & 0.662 & 0.500 & \textbf{0.439} \\
 & ImageNN      & 0.182 & \textbf{0.426} & 0.500 & 0.527 \\
 & EF-NN        & \textbf{0.062} & 0.695 & 0.500 & 0.495 \\
 & IF-NN        & 0.816 & 0.731 & 0.500 & 0.571 \\
 & MC-R         & 0.087 & 0.658 & 0.500 & 0.472 \\
 & MPS          & 0.079 & 0.662 & 0.500 & \textbf{0.439} \\
\midrule
\multirow{6}{*}{Jobs}
 & TabularNN    & 0.008 & 0.657 & 0.216 & \textbf{0.034} \\
 & TextNN       & 0.005 & 0.688 & 0.241 & 0.113 \\
 & EF-NN        & 0.017 & 0.725 & 0.216 & 0.035 \\
 & IF-NN        & 0.113 & 0.906 & 0.241 & 0.098 \\
 & MC-R         & \textbf{0.000} & \textbf{0.520} & \textbf{0.137} & 0.040 \\
 & MPS          & 0.026 & 0.740 & 0.213 & 0.039 \\
\bottomrule
\end{tabular}}
\end{table}

Table~\ref{tab:generator_perdataset} averages across all three fusion strategies; however, the fusion strategy of the explained classifier is itself a source of variance. Its effect is concentrated on outcome and is dataset-dependent: on Long COVID Tweets, the choice of fusion strategy shifts mean outcome by up to 0.58, while on Sepsis and Jobs the range is below 0.05. Proximity, sparsity, and plausibility are largely stable across fusion strategies.
\begin{comment}
% ----------------------------------------------------------
\subsection{Sensitivity to Fusion Strategy}
\label{sec:fusion_sensitivity}
% ----------------------------------------------------------
 
Fig.~\ref{fig:fusion_stability} shows per-generator metric values broken down by fusion strategy for the Sepsis dataset.
 
\begin{figure}
    \centering
    \includegraphics[width=0.99\linewidth]{figures/stability_sepsis.pdf}
    \caption{Generator quality on Sepsis broken down by fusion strategy
    (early / intermediate / late), at $k = 20$.}
    \label{fig:fusion_stability}
\end{figure}
 
The fusion strategy of the explained classifier has a pronounced effect on outcome but leaves proximity, sparsity, and plausibility largely unchanged. On Hateful Memes, intermediate fusion raises the mean outcome to 0.279 against 0.056-0.061 for early and late fusion. On Long COVID Tweets, the early fusion classifier is the hardest target (mean outcome 0.680), compared to 0.127 for intermediate and 0.101 for late fusion. On Sepsis and Jobs the fusion strategy has comparatively little effect: the outcome range across strategies is 0.042 on Sepsis and 0.020 on Jobs.
\end{comment}
% ----------------------------------------------------------
\subsection{Effect of \texorpdfstring{$k$}{k} on Counterfactual Quality}
\label{sec:k_ablation}
% ----------------------------------------------------------
 
\begin{figure}
    \centering
    \includegraphics[width=0.9\linewidth]{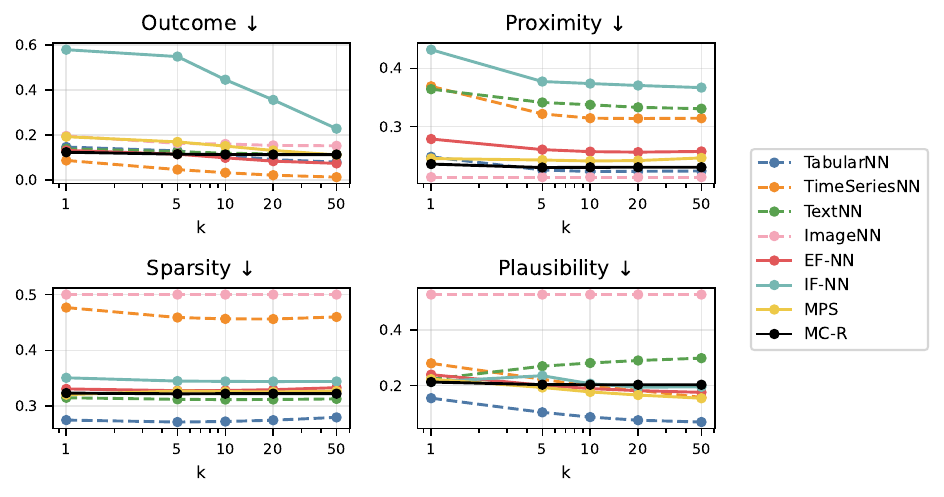}
    \caption{Effect of $k$ on counterfactual quality, averaged across all datasets and fusion strategies. Unimodal generators are shown with dashed lines, multimodal with solid lines.}
    \label{fig:k_ablation}
\end{figure}

Fig.~\ref{fig:k_ablation} shows how each metric evolves as $k$ increases from 1 to 50, averaged across all datasets and fusion strategies.
We observe that outcome improves monotonically with $k$ for nearly all generators, with the largest gains for TabularNN (0.178 to 0.049), EF-NN (0.170 to 0.052), MPS (0.246 to 0.095), and IF-NN (0.512 to 0.198), which shows the largest absolute improvement, yet remains the worst generator even at $k=50$. 
MC-R's outcome plateaus almost immediately after $k=5$ (0.116 to 0.115 at $k=50$): adding new instances to the consensus set does not substantially improve the outcome. Plausibility improves with $k$ for structured modalities as the progressive median prototype moves toward the centre of the target-class neighbourhood, but degrades for TextNN, since larger $j$ draws from more distant training instances. ImageNN plausibility is flat (0.527 at all $k$), as every image candidate is a real training instance. Sparsity is nearly flat across $k$ for all generators.

% ----------------------------------------------------------
%  SECTION 6 -- DISCUSSION
% ----------------------------------------------------------
\section{Discussion}
\label{sec:discussion}

\noindent\textbf{Unimodal retrieval.} 
Unimodal generators achieve strong proximity and plausibility because they retrieve complete, coherent training instances: cross-modal coherence is preserved by construction. On Sepsis, TimeSeriesNN achieves near-perfect outcome (0.021) because the temporal clinical trajectory dominates the classifier's decision boundary, consistent with prior findings on this dataset~\cite{mondrejevski2024masicu}; the retrieved instance's static features jointly reinforce the flip. On imbalanced datasets (Sepsis, Jobs), performance is partly aided by the large majority-class retrieval pool; practitioners should expect degradation when the target class is the minority.
Conversely, on Hateful Memes, TextNN (0.070) substantially outperforms ImageNN (0.182), suggesting that the text modality dominates the classifier's decision boundary for meme classification, consistent with prior findings on this benchmark~\cite{kiela2020hateful}.
\smallskip

\noindent\textbf{Latent-space retrieval.}
IF-NN consistently fails to produce valid counterfactuals across all datasets and fusion strategies, even at $k=50$. The penultimate layer of a deep classifier is optimised for discriminative separation rather than within-class clustering, so neighbours in that space are not reliably from the target class. This limitation is specific to current training objectives; architectures that explicitly optimise for clustered latent representations may overcome it.
\smallskip

\noindent\textbf{Classifier quality.}
Counterfactual quality is partly a function of classifier quality: a well-separating classifier assigns distinct predictions to distinct classes, so retrieved target-class instances are more likely to flip the prediction. This is consistent with the results: on Jobs, where all classifiers achieve F1 above 0.88, every generator except IF-NN reaches near-zero outcome, whereas on Hateful Memes, which has the weakest classifiers in the benchmark (F1 0.669-0.713), outcome is substantially higher across all generators
\smallskip

\noindent\textbf{MC-R as the recommended multimodal strategy.}
MC-R is the strongest multimodal generator overall, achieving the best or joint-best outcome on three of four datasets and perfect outcome on Real or Fake Jobs (0.000). On Jobs, MC-R also achieves the lowest sparsity in the entire benchmark (0.137), indicating that the intersection mechanism identifies target-class instances that differ from the query in very few features, a desirable property for actionable explanations in fraud detection. Its intersection mechanism identifies instances jointly close to the query across all modalities, grounding counterfactuals in the joint training distribution in a way that MPS, which synthesises per-modality prototypes independently, does not guarantee. Its outcome saturates at $k=5$, making it efficient in practice. The choice of $k$ also governs a trade-off between coherence and diversity~\cite{mothilal2020explaining}: at $k=1$, the proposed candidate is a real, coherent training instance; increasing $k$ yields a larger and more diverse candidate set, with the first candidate always retaining full cross-modal coherence.

\begin{comment}
\noindent\textbf{Effect of $k$.}
Outcome improves monotonically with $k$ for most generators at no sparsity cost. MC-R saturates at $k=5$; text modalities degrade in plausibility with increasing $k$ since larger $j$ draws from more distant instances. MPS benefits most consistently across all objectives simultaneously.

\smallskip
\noindent\textbf{Limitations.}
IF-NN requires a deep neural classifier with accessible activations. Sparsity and proximity for text and image are measured in the embedding space, which is less directly interpretable than feature-level metrics; practitioners can substitute any discrete-space distance without modifying the pipeline. No human evaluation of counterfactual quality has been conducted. The Long COVID Tweets evaluation covers only the sadness$\to$joy direction.
\end{comment}

% ----------------------------------------------------------
%  SECTION 7 -- CONCLUSION
% ----------------------------------------------------------
\section{Conclusion}
\label{sec:conclusion}

We presented CRANE, a post-hoc, model-agnostic framework for nearest-neighbour counterfactual generation for multimodal classifiers, supporting arbitrary modality combinations. Its extensibility to new modalities makes it naturally applicable to domains, such as clinical decision support, where normative frameworks require ML tools to be explainable~\cite{freyer2024ethical}. Across four datasets and three fusion strategies, MC-R emerges as the strongest multimodal strategy, combining the strongest outcome scores in the benchmark, joint-distribution grounding, and computational efficiency. %Unimodal generators excel when a single modality dominates the decision boundary. IF-NN is a cautionary result: latent-space retrieval using the classifier's own representation reliably fails, a consequence of optimising for discrimination rather than clustering. Proximity, sparsity, and plausibility are stable across classifier families, supporting the model-agnostic claim.
Future work includes comparative evaluation against the baselines described in Section~\ref{sec:related}, human evaluation of counterfactual quality with domain experts, extension to additional clinical datasets, and exploration of generative counterfactual strategies within the CRANE evaluation protocol.

\bibliographystyle{IEEEtran}
\bibliography{bibliography}
\end{document}